%% file: main.tex
\documentclass[10pt,twocolumn]{ICCAS}
 
\usepackage{diagbox}
\usepackage{multirow}
\usepackage{url}  

\begin{document}

\title{Impact of Dataset Composition on Embedded Real-Time UAV Wildfire Detection Using Compact YOLO Models}

\author{Eduardo de los Santos${}^{1}$, Andre S. Kelbouscas${}^{1}$, Ricardo B. Grando${}^{1*}$ and Bruna V. Guterres${}^{1}$}

\affils{ ${}^{1}$Robotics and AI Lab, Technological University of Uruguay, \\
Rivera, Uruguay (ricardo.bedin@utec.edu.uy){\small${}^{*}$ Corresponding author}}

% \thanks{ \noindent
%   This paper is supported by my funding agencies.
%  }

\input{sessions/1_abstract}


\maketitle

% %-----------------------------------------------------------------------

\input{sessions/2_intro}
\input{sessions/3_related_works}
\input{sessions/4_methodology}

\input{sessions/5_results}
\input{sessions/6_conclusion}

\bibliographystyle{./bibliography/IEEEtran}
\bibliography{./bibliography/IEEEabrv,./bibliography/main}

\end{document}

%% file: sessions/1_abstract.tex
\abstract{
   The development of vision-based wildfire detection systems for unmanned aerial vehicles is constrained by the limited availability of diverse real-world training images. This paper investigates the impact of dataset composition on embedded real-time UAV wildfire detection using compact YOLO models as a controlled validation family. Four training configurations were evaluated: real non-augmented, real augmented, hybrid non-augmented, and hybrid augmented, where the hybrid sets combine real wildfire images with AI-generated samples. The objective is to determine whether synthetic data mixing and image augmentation improve practical detection performance under resource-constrained deployment conditions. Experimental results show that the best overall operating point was obtained with the real non-augmented dataset, which achieved the strongest balance between recall and mean average precision for UAV-based wildfire detection. The results also show that neither hybridization with synthetic data nor augmentation produced a better final deployment choice. These findings suggest that, for embedded UAV wildfire detection, dataset realism and domain alignment are more valuable than increasing training set size through synthetic expansion.
}

%% file: sessions/2_intro.tex
\section{Introduction}

Wildfires remain a major environmental and operational problem. Early detection is still difficult in large and remote areas, where fixed infrastructure is sparse and continuous human observation is costly. In this context, unmanned aerial vehicles have become an attractive option because they offer flexible coverage, fast deployment, lower operational cost, and reduced exposure of personnel to hazardous environments \cite{Bouguettaya2022,Boroujeni2024,BailonRuiz2022}.

Recent progress in deep learning has made image based smoke and fire detection more feasible for UAV platforms. In particular, detectors from the YOLO family are attractive for real-time applications because object localization and classification are performed in a single forward pass \cite{Redmon2016}. The widespread use of compact YOLO models in edge vision applications has made them a practical choice for embedded deployment, especially when model compactness and inference speed must be balanced against detection quality \cite{Khanam2024,Mukhiddinov2022,Yang2023}. Recent UAV wildfire studies have also reported onboard or edge based perception pipelines using NVIDIA Jetson hardware and YOLO based detectors, which confirms the relevance of this design space \cite{Shamta2024,Pesonen2025}.

At the same time, data remain a limiting factor. Publicly available wildfire datasets are still relatively scarce, especially from aerial viewpoints and with annotations suitable for object detection or segmentation \cite{PesonenDataset2025}. This is a recurring issue in wildfire vision research and one reason why synthetic data and hybrid training sets are often considered as a way to increase data diversity \cite{Mukhiddinov2022,Richter2016}. However, the practical value of synthetic expansion must be validated for the target deployment setting rather than assumed a priori.

In this work, a wildfire dataset was assembled from real UAV imagery, public datasets, images provided by the Uruguayan Air Force remote sensing service, and AI-generated synthetic images. Four compact YOLO models were trained and evaluated under four dataset configurations: real non-augmented, real augmented, hybrid non-augmented, and hybrid augmented. Rather than focusing on detector architecture itself, this paper investigates whether changes in dataset composition improve embedded real-time UAV wildfire detection under resource-constrained conditions. The compact YOLO models are therefore used as a controlled validation family, while the primary contribution of the study lies in the comparative analysis of how real, augmented, and hybrid data affect practical deployment performance.

\begin{figure*}[t]
    \centering
    \includegraphics[width=0.99\textwidth]{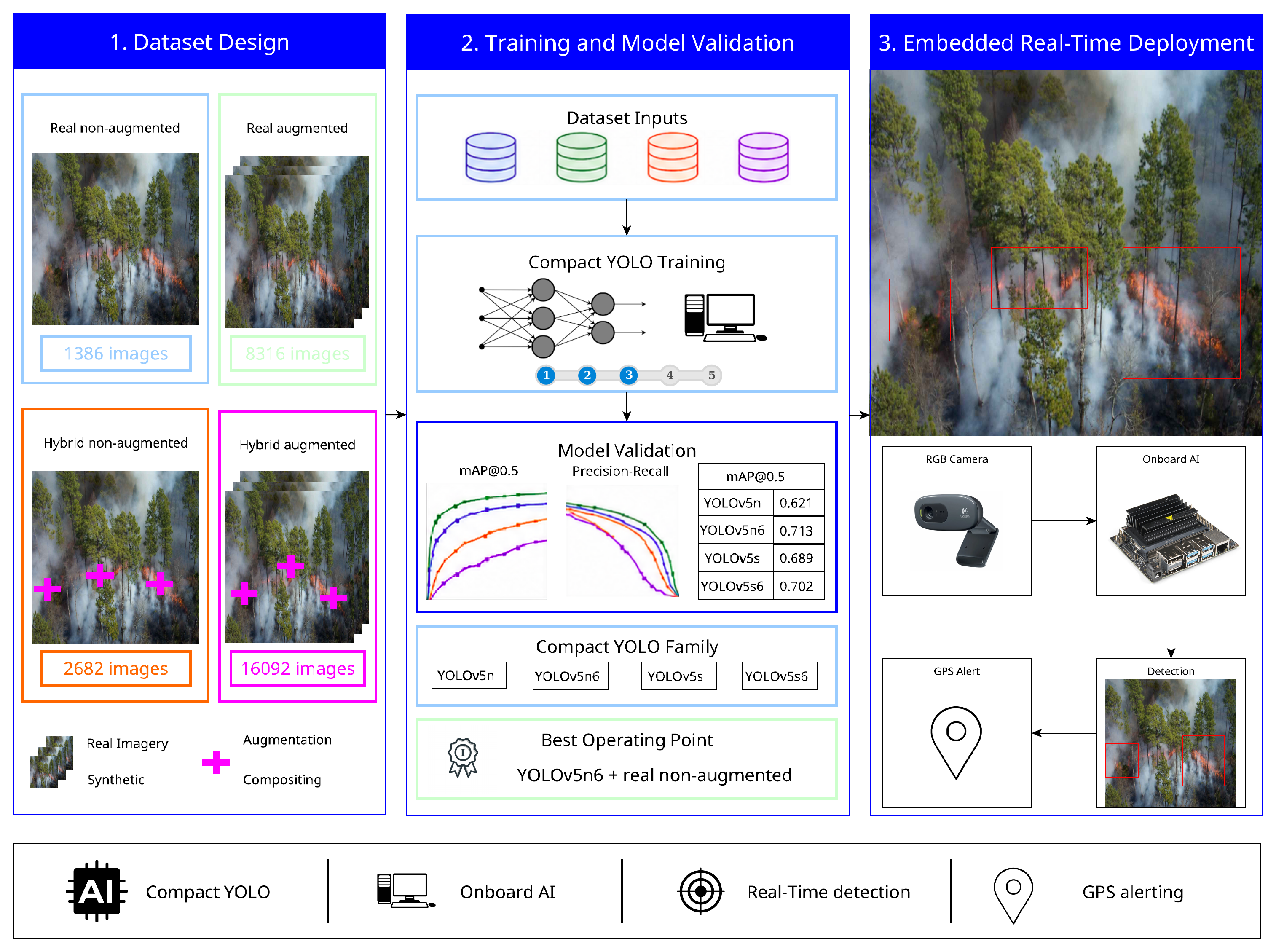}
    \caption{Overview of the proposed embedded real-time UAV wildfire detection framework. The figure summarizes the central UAV platform, the four dataset configurations evaluated in the study, and the compact YOLO models used as a controlled validation family for analyzing the effect of dataset composition on detection performance.}
    \label{fig:overview_framework}
\end{figure*}

Figure~\ref{fig:overview_framework} provides an overview of the proposed framework. The UAV platform is shown at the center as the embedded deployment target, while the four surrounding quadrants represent the dataset configurations evaluated in this study. The lower part of the figure summarizes the compact YOLO models used as a controlled validation family and the evaluation metrics considered in the experimental analysis. 
% Complementary material\footnote{Video available at: \url{https://youtu.be/PSGrWwR8Rug?si=8eH41TkTl3W5whPM}} shows our vehicle performing real-time detection.

The main contributions of this paper are as follows:
\begin{itemize}
    \item A dataset centered study is presented for embedded real-time UAV wildfire detection using four training configurations based on real, augmented, hybrid, and hybrid augmented data.
    \item Compact YOLO models are used as controlled evaluators under a common real-time object detection framework suitable for embedded deployment.
    \item It is shown that the best practical deployment tradeoff was obtained with the real non augmented dataset, indicating that dataset realism was more valuable than synthetic expansion in the evaluated scenario.
\end{itemize}

%% file: sessions/3_related_works.tex
\section{Related Work}

UAV based wildfire monitoring has been studied from both system and perception perspectives. Survey papers show that UAVs are increasingly used for detection, monitoring, and management tasks because they can provide rapid and low cost observations in areas where traditional methods are incomplete or delayed \cite{Bouguettaya2022,Boroujeni2024}. In addition, real time UAV monitoring frameworks have been developed for wildfire observation with planning and control components, showing that timely aerial information can improve operational awareness during fire events \cite{BailonRuiz2022}.

Within vision based wildfire detection, YOLO based models have become common because they offer a favorable tradeoff between inference speed and detection quality. YOLO introduced single stage real time object detection as a unified regression problem \cite{Redmon2016}. More recent works have documented the practical value of YOLOv5 for edge and embedded use, including its architecture, training workflow, and deployment characteristics \cite{Khanam2024}. In wildfire specific applications, Mukhiddinov et al. proposed an optimized YOLOv5 system for wildfire smoke detection from UAV images and emphasized the need for more reliable datasets \cite{Mukhiddinov2022}. Yang et al. proposed a lightweight improved YOLOv5 model for forest smoke and fire detection and reported reductions in model size and FLOPs for resource constrained deployment \cite{Yang2023}.

Embedded UAV fire detection systems have also been reported. Shamta and Demir developed a UAV surveillance system with onboard NVIDIA Jetson Nano processing and compared YOLOv5 and YOLOv8 for object detection, together with a ground station interface for fire related information \cite{Shamta2024}. More recent work by Pesonen et al. has shown that real time wildfire perception can be achieved with onboard computation on UAV carried hardware, although training data limitations remain a major issue \cite{Pesonen2025}.

Also, recent published datasets confirm that open access UAV wildfire data are still limited, particularly for smoke focused annotations and detailed supervision \cite{PesonenDataset2025}. In computer vision more broadly, synthetic data have often been used to supplement real data and improve model training when manual annotation is expensive \cite{Richter2016}. However, whether such gains transfer to embedded UAV wildfire detection remains an empirical question.

Based on this literature, the gap addressed here is not the proposal of a new detector. Instead, the gap lies in the controlled comparison of dataset composition strategies for embedded real time UAV wildfire detection, using a compact YOLO validation family under the same experimental protocol.

\begin{figure}[t]
    \centering
    \includegraphics[width=0.48\textwidth]{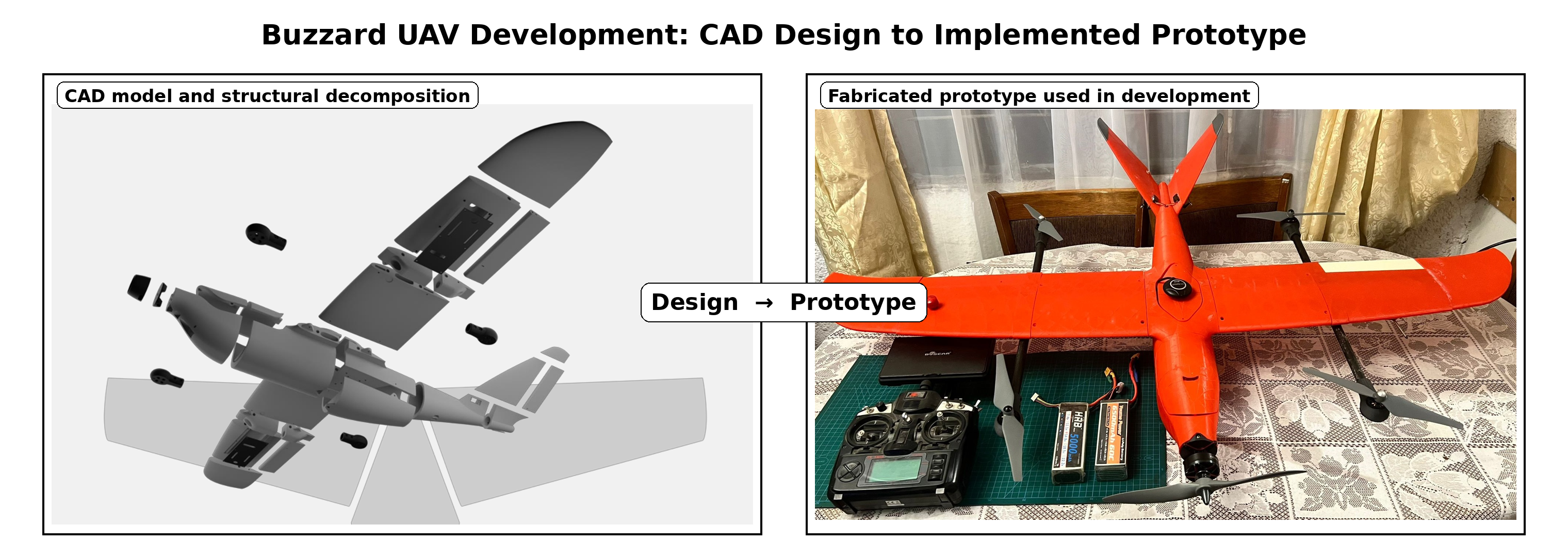}
    \caption{Development path of the embedded wildfire monitoring UAV. The left panel shows the Buzzard CAD model and structural decomposition used during design and integration planning, while the right panel shows the fabricated prototype adapted for the real-time wildfire detection payload. }
    \label{fig:buzzard_design_to_prototype}
\end{figure}

%% file: sessions/4_methodology.tex
\section{System Overview and Experimental Design}

\subsection{Embedded UAV Detection System}

The proposed wildfire monitoring platform was implemented on a custom 3D printed Buzzard UAV, selected because it combines a lightweight airframe, vertical takeoff capability, and an internal cargo bay suitable for embedded sensing and onboard computation. The vehicle has a wingspan of 1.3~m, a maximum takeoff weight in the range of 2.5--3.5~kg, a cruise speed of 50--75~km/h, and an operational range of approximately 45--65~km. Its configuration follows a 4+1 propulsion layout, with four vertical lift motors for takeoff and landing and one nose mounted propulsion motor for horizontal flight. This architecture enables VTOL operation while preserving the endurance advantages of fixed wing forward flight, which is beneficial for real-time wildfire monitoring over extended areas.

% The airframe was manufactured using fused filament fabrication. PLA was used for the primary structural elements, while carbon fiber reinforced materials were employed for motor mounts and electronics supports in order to improve stiffness and mechanical robustness. The base design was adapted from the Buzzard vehicle of Tytan Dynamics and then modified for the present application. In particular, the detachable cargo bay was redesigned to accommodate the embedded wildfire detection payload, including the NVIDIA Jetson Nano, the monocular RGB camera, and the communication electronics required to issue alerts and transmit approximate GPS coordinates.

% The propulsion system combines four EMAX MT2216 II 810KV motors for vertical lift with one T-MOTOR AM480 600KV nose motor for forward thrust. During takeoff and landing, the vehicle behaves similarly to a multirotor platform, while in cruise it transitions to horizontal flight, which improves endurance relative to conventional quadrotor systems. The UAV is powered by two LiPo batteries dedicated to flight functions, rated at 6500~mAh and 5000~mAh, respectively, and connected through a power distribution board.

A key feature of the system is the integration of the embedded wildfire detection unit in the cargo bay. The onboard processor is an NVIDIA Jetson Nano 4~GB, chosen because it provides GPU acceleration and compatibility with deep learning frameworks under reduced size, weight, and power constraints. In the implemented platform, the Jetson Nano is paired with an RGB camera for onboard image acquisition and inference. The embedded unit is powered independently through a 10{,}000~mAh power bank, which was modified to reduce unnecessary casing weight. This split power architecture isolates the payload electronics from the flight subsystem and allows the detector to be activated only when the UAV reaches the area of interest, thus improving energy management during operation.

% The flight stack is built around a Matek F405-WTE autopilot running ArduPilot. The platform integrates an accelerometer, gyroscope, GPS receiver, pitot tube, landing radar, and altimeter, which support both manual and autonomous flight, including waypoint navigation, altitude control, and automatic landing. The sensing and avionics components were distributed to preserve center of gravity and flight stability after payload integration.

% Communication is divided into three independent channels. Telemetry is transmitted through a 900~MHz link, which provides GPS position, altitude, speed, battery information, and other flight variables to the ground station. Manual piloting is achieved with a 2.4~GHz FlySky radio control interface. Real-time video transmission is provided through a 5.8~GHz analog link based on the Rush Tank Solo transmitter. This architecture is consistent with recent embedded UAV wildfire monitoring systems, where local processing is preferred in order to reduce dependence on cloud connectivity and to support real-time detection in operational conditions \cite{Shamta2024,Pesonen2025,BailonRuiz2022}.

From a manufacturing perspective, the UAV was developed through design review, 3D printing, structural assembly, payload integration, and flight validation. The parts were printed using an Ender-3 Max Neo printer with PLA, PLA-LW, and ePA-CF materials, with an approximate fabrication time of 150 hours per vehicle. Two airframes were produced for redundancy during development, although only one was required in the final implementation. Figure~\ref{fig:buzzard_design_to_prototype} illustrates the progression from the conceptual CAD decomposition of the Buzzard airframe to the fabricated prototype used for embedded wildfire detection experiments.

\subsection{Dataset Sources}

The image pool used in this study was assembled from both real and synthetic wildfire imagery. The real data comprised 1386 images collected from three sources: 447 images captured with our UAV, 857 images obtained from public datasets, and 82 images provided by the remote sensing service of the Uruguayan Air Force. These sources contributed variability in viewpoint, altitude, scene composition, and fire appearance, which is important for detector robustness in real aerial operation.

In addition, 1296 synthetic wildfire images were generated. These synthetic images were curated manually to retain visually plausible wildfire scenes. The use of synthetic imagery was motivated by the limited availability and high acquisition cost of real aerial wildfire images, a challenge also recognized in the broader wildfire monitoring literature \cite{Bouguettaya2022,Mukhiddinov2022,PesonenDataset2025}. Synthetic image generation was therefore treated as a practical mechanism for expanding data diversity, while still requiring validation against real deployment conditions.

\subsection{Dataset Configurations}

Four dataset configurations were defined in order to study the effect of training data composition on embedded real-time UAV wildfire detection:

\begin{itemize}
    \item \textbf{Real non augmented}: 1386 real wildfire images.
    \item \textbf{Real augmented}: the same real image set after augmentation, expanded to 8316 images.
    \item \textbf{Hybrid non augmented}: 1386 real images combined with 1296 synthetic images, totaling 2682 images.
    \item \textbf{Hybrid augmented}: augmented real data combined with augmented synthetic data, totaling 16092 images.
\end{itemize}

All datasets were divided into 70\% for training, 20\% for validation, and 10\% for testing. This design makes dataset composition the central experimental variable of the paper. Rather than focusing on architectural novelty, the study evaluates whether real data augmentation and synthetic hybridization improve detection performance under embedded real-time deployment constraints.

\subsection{Annotation and Augmentation Strategy}

The annotation was performed with Roboflow, and the labeling protocol was kept consistent across all sources in order to reduce annotation bias. A review stage was carried out to identify and correct possible errors before training. The augmentation pipeline applied to the real and hybrid datasets included random crop followed by resize to $640 \times 640$, horizontal and vertical flips, random rotations up to $\pm 30^\circ$, random brightness and contrast variation, and blur simulation. These transformations were introduced to test whether increased visual diversity would improve detector robustness under real-world UAV imaging conditions.

\subsection{Compact YOLO Models as Validation Family}

In the present study, the compact YOLO validation family was instantiated by four YOLOv5 variants: YOLOv5n, YOLOv5n6, YOLOv5s, and YOLOv5s6. These models were selected because compact YOLO variants are appropriate for embedded real-time detection, where low latency and reduced computational cost must be balanced against recall and localization quality. In YOLOv5, the \textit{n} and \textit{n6} variants are designed for strongly resource-constrained platforms, while the \textit{s} and \textit{s6} variants offer a slightly larger model capacity while remaining suitable for real-time use \cite{Khanam2024,Yang2023}.

The compact YOLO models were not treated as competing contributions; instead, they were used to evaluate how detector performance changes under different dataset compositions. Instead, they serve as a controlled validation family for testing whether the conclusions about dataset composition remain stable across closely related detector capacities suitable for the NVIDIA Jetson Nano class of embedded hardware.

\subsection{Training and Evaluation Protocol}

All models were trained under the same protocol for 1000 epochs with batch size 32 and patience 100. Training was performed on a workstation equipped with an Intel Core i7-12700K CPU, 40~GB of RAM, and an NVIDIA GeForce RTX 4060 Ti GPU with 8~GB of memory. Depending on the selected model and dataset configuration, training time ranged from approximately 10 to 30 hours.

Performance was evaluated using Precision, Recall, mAP@0.5, mAP@0.5:0.95, and Box Loss. These metrics were selected because wildfire monitoring requires a balance between correct positive detection, localization quality, and practical deployment behavior. In particular, recall and mAP were emphasized in the final model interpretation, since missing a real wildfire event is generally more costly than accepting a limited number of false alarms in an early warning system.

This experimental design allows the impact of dataset composition to be isolated under a consistent family of embedded real-time detectors. As a result, the following sections interpret detector behavior not primarily as a model ranking problem, but as a controlled study of how real, augmented, and hybrid data affect practical UAV wildfire detection performance. Accordingly, the term real-time in this paper refers to deployment-oriented onboard operation on an embedded UAV platform, supported by practical flight tests and model compactness considerations rather than by a dedicated latency benchmarking study.

%% file: sessions/5_results.tex
\section{Results and Discussion}

Table~\ref{tab:all_models_all_datasets} presents the complete performance comparison of all compact YOLO models under the four dataset configurations considered in this study: real non-augmented, real augmented, hybrid non-augmented, and hybrid augmented. Unlike a reduced summary containing only the best model per dataset, this merged view makes it possible to analyze how each compact detector responds to changes in dataset composition, including real data only, data augmentation, synthetic hybridization, and the combination of both. This perspective is important because the main contribution of the paper lies in understanding how training data design affects embedded real-time UAV wildfire detection rather than in proposing a new detector architecture.

\begin{table*}[t]
\centering
\caption{Comprehensive comparison of all compact YOLO models under the four dataset configurations: real non augmented, real augmented, hybrid non augmented, and hybrid augmented. Bold values indicate the best result within each dataset block.}
\label{tab:all_models_all_datasets}
\small
\setlength{\tabcolsep}{14pt}
\begin{tabular}{llccccc}
\hline
Dataset & Model & Precision & Recall & mAP@0.5 & mAP@0.5:0.95 & Box Loss \\
\hline
\multirow{4}{*}{Real non augmented}
& YOLOv5n  & 0.786490 & 0.702290 & 0.696410 & 0.310220 & 0.056728 \\
& YOLOv5n6 & 0.793000 & \textbf{0.720470} & \textbf{0.729490} & \textbf{0.313290} & 0.044101 \\
& YOLOv5s  & 0.813110 & 0.718690 & 0.709570 & 0.312170 & 0.052995 \\
& YOLOv5s6 & \textbf{0.813250} & 0.707340 & 0.704460 & 0.308920 & \textbf{0.037963} \\
\hline
\multirow{4}{*}{Real augmented}
& YOLOv5n  & 0.756930 & 0.672680 & 0.670800 & 0.294180 & 0.062370 \\
& YOLOv5n6 & 0.733490 & 0.684870 & 0.669710 & 0.289830 & 0.051225 \\
& YOLOv5s  & 0.753300 & 0.670360 & 0.680800 & 0.287830 & 0.085911 \\
& YOLOv5s6 & \textbf{0.798450} & \textbf{0.710320} & \textbf{0.685670} & \textbf{0.301230} & \textbf{0.039870} \\
\hline
\multirow{4}{*}{Hybrid non augmented}
& YOLOv5n  & \textbf{0.77394} & 0.62690 & \textbf{0.69852} & \textbf{0.28426} & 0.092927 \\
& YOLOv5n6 & 0.67358 & 0.60156 & 0.62122 & 0.26830 & 0.074508 \\
& YOLOv5s  & 0.71916 & \textbf{0.66409} & 0.66675 & 0.27300 & 0.092040 \\
& YOLOv5s6 & 0.72731 & 0.57787 & 0.61312 & 0.28255 & \textbf{0.071631} \\
\hline
\multirow{4}{*}{Hybrid augmented}
& YOLOv5n  & 0.68531 & 0.62640 & 0.65412 & 0.26674 & 0.08394 \\
& YOLOv5n6 & 0.68410 & 0.60391 & 0.65671 & 0.26810 & 0.07845 \\
& YOLOv5s  & 0.70023 & \textbf{0.64732} & 0.66850 & 0.27390 & 0.07389 \\
& YOLOv5s6 & \textbf{0.70345} & 0.63011 & \textbf{0.67212} & \textbf{0.28145} & \textbf{0.07162} \\
\hline
\end{tabular}
\end{table*}

\subsection{Performance Across Dataset Configurations}

The results obtained with the real non augmented dataset show the strongest overall detection behavior among all evaluated configurations. In this setting, YOLOv5n6 achieved the highest recall, the highest mAP@0.5, and the highest mAP@0.5:0.95, while YOLOv5s6 achieved the highest precision and the lowest Box Loss. This indicates that the real non augmented dataset provided the best balance between detection robustness and localization quality. From a deployment perspective, this result is especially relevant because recall and mAP are more informative than precision alone in early wildfire detection, where missing a true event may be more costly than issuing a limited number of false alarms.

The real augmented dataset remained competitive, but it did not surpass the real non augmented setting. In this case, YOLOv5s6 dominated all reported metrics, including precision, recall, mAP@0.5, mAP@0.5:0.95, and Box Loss. This suggests that augmentation improved the internal consistency of the training process for the larger compact variant, but it did not lead to a better overall operating point than that obtained from real images without augmentation. Therefore, increasing the number of training samples through geometric and photometric transformations alone was not sufficient to improve the best final deployment choice.

The hybrid non augmented dataset produced a more fragmented performance pattern. YOLOv5n achieved the highest precision and the best mAP values within this dataset block, whereas YOLOv5s achieved the highest recall and YOLOv5s6 achieved the lowest Box Loss. These results suggest that adding synthetic imagery can still yield useful detectors, but the gains were less consistent than in the real only setting. In particular, the hybrid non augmented dataset did not preserve the stronger recall and mAP behavior observed with the real non augmented configuration, which indicates that synthetic expansion did not fully match the visual statistics of the real aerial wildfire scenes targeted for deployment.

A similar trend appears in the hybrid augmented dataset. In this case, YOLOv5s6 achieved the highest precision, the best mAP@0.5, the best mAP@0.5:0.95, and the lowest Box Loss, while YOLOv5s achieved the highest recall. Although the hybrid augmented configuration improved some metrics relative to the hybrid non augmented one, it still did not exceed the best values obtained with the real non augmented dataset. This result indicates that combining synthetic data with augmentation increased diversity, but did not produce the strongest final operational behavior. In the present application, model selection prioritizes recall and mAP over precision alone, since early wildfire detection penalizes missed events more severely than a limited number of false positives.

\subsection{Impact of Dataset Composition}

When the results are analyzed from the perspective of dataset composition rather than isolated model ranking, a clear pattern emerges. The real non augmented dataset yielded the highest mAP@0.5 among the best models from each configuration, and it also produced the highest recall. In contrast, the best hybrid configurations showed lower recall and lower mAP, even when precision remained competitive. This suggests that, for embedded real-time UAV wildfire detection, preserving the original appearance and domain characteristics of real imagery was more beneficial than increasing dataset size through synthetic expansion.

This conclusion is also consistent with the dataset construction process itself. The complete image pool combined drone captures, public datasets, imagery from the Uruguayan Air Force remote sensing service, and AI generated synthetic samples. While the synthetic images contributed diversity, the dataset analysis already indicated that they still differed from real imagery in texture and scene naturalness. The present results suggest that these visual differences matter in practice, especially when the detector must generalize to real aerial video captured under operational conditions. In other words, more training data did not necessarily mean better training data for the target deployment domain.

Another important observation is that the different compact YOLO variants exhibited complementary strengths. YOLOv5n6 was particularly strong in recall and mAP when trained on real non augmented data, which made it the most suitable model for practical wildfire detection in flight. In contrast, YOLOv5s6 repeatedly achieved the lowest Box Loss and often the highest precision, especially in the augmented settings, which indicates better localization accuracy and more conservative detection behavior. This distinction is useful because it shows that compact YOLO models can serve as a meaningful validation family for dataset studies: model scale affects the balance between detection sensitivity and localization precision, but the dominant trend in this study still came from the dataset composition.

Taken together, these results support the main claim of the paper. The most important factor in the evaluated embedded UAV setting was not the choice between closely related compact YOLO variants, but the composition of the training data itself. The best practical outcome was obtained when the detector was trained on real, non augmented imagery, which preserved the domain characteristics of the final operating environment.

\subsection{Real-World UAV Validation}

The real-world UAV tests further support the quantitative interpretation of the dataset comparison. Videos captured from the platform over areas of interest showed that the selected model was capable of detecting wildfire events effectively during real-time operation, even under motion and viewpoint variation. This is important because it demonstrates that the configuration identified as best through the offline quantitative evaluation was also useful under realistic aerial conditions.

From an application perspective, this result strengthens the case for using compact YOLO models on embedded UAV platforms when early warning capability is required. The real non-augmented dataset preserved the visual characteristics of the target environment more faithfully, and the detector trained on it transferred more effectively to aerial video. This makes the selected configuration more relevant for practical wildfire monitoring missions in which onboard processing and rapid geographic reporting are required.

\subsection{Discussion}

From a broader perspective, the findings indicate that dataset realism should be treated as a first-order design variable in embedded wildfire detection systems. Synthetic images and data augmentation remain valuable tools, especially when real annotated imagery is scarce. However, the present results show that their usefulness depends on how well they preserve the visual characteristics of the target deployment domain. For this reason, future work should focus not only on enlarging training sets, but also on improving the fidelity and domain alignment of synthetic or augmented wildfire imagery.

These results also reinforce the practical relevance of compact YOLO models for embedded aerial perception. The evaluated models were lightweight enough for resource-constrained deployment, yet they still revealed clear differences in how dataset composition affects real-time detection behavior. In this sense, the compact YOLO family was appropriate not because it introduced architectural novelty, but because it provided a stable experimental basis for identifying how training data design affects embedded UAV wildfire detection performance.

%% file: sessions/6_conclusion.tex
\vspace{-5mm}
\section{Conclusion}

This paper presented a dataset-centered study of embedded real-time UAV wildfire detection. Rather than focusing on detector architecture, the work evaluated how changes in training dataset composition affect detection performance under resource-constrained deployment conditions, using compact YOLO models as a controlled validation family.

The results showed that the best practical operating point was obtained with YOLOv5n6 trained on the real non-augmented dataset. In contrast, neither augmentation nor hybridization with AI-generated imagery produced a better final deployment recommendation. These findings indicate that, in the evaluated wildfire detection scenario, dataset realism was more valuable than increasing training set size through synthetic expansion.

Future work should investigate domain-adapted synthetic image generation, broader cross-dataset validation, and larger collections of real aerial wildfire imagery acquired under different weather, altitude, and illumination conditions. Additional onboard benchmarking with newer embedded platforms may also help determine under which conditions hybrid data strategies become beneficial.

\vspace{-5mm}
\section{Acknowledgements}

The authors of this work would like to thank Professor PhD David Saldaña of Lehigh University for his technical support in this work. We also would like to thank the Technological University of Uruguay and the Laboratory of Robotics and AI for the support in this work.